\documentclass{article}

\RequirePackage{silence}
\usepackage[final]{staix_2026}         

\usepackage[utf8]{inputenc}
\usepackage[T1]{fontenc}
\usepackage{hyperref}
\usepackage{url}
\usepackage{booktabs}
\usepackage{amsfonts}
\usepackage{nicefrac}
\usepackage{microtype}
\usepackage{xcolor}
\usepackage{graphicx}
\usepackage{amsmath,amssymb,amsthm}
\usepackage{multirow}
\usepackage{array}
\usepackage{tabularx}
\usepackage{comment}
\usepackage{bm}

\newcommand{\red}{\color{red}}
\newcommand{\blue}{\color{blue}}
\newcommand{\gray}{\color{gray}}

\newcommand{\bbR}{\mathbb{R}}

\newcommand{\cD}{{\cal D}}
\newcommand{\cF}{{\cal F}}
\newcommand{\cL}{{\cal L}}
\newcommand{\cN}{{\cal N}}
\newcommand{\cP}{{\cal P}}
\newcommand{\cS}{{\cal S}}
\newcommand{\cY}{{\cal Y}}

\newcommand{\best}[1]{\textbf{\textcolor{red}{#1}}}
\newcommand{\second}[1]{\textcolor{blue}{\underline{#1}}}

\newcolumntype{Y}{>{\centering\arraybackslash}X}

\title{Adaptive Kernel Density Estimation with Pre-training}

\author{%
  Ruitong Zhang \\
  Department of Statistics and Data Science \\
  Tsinghua University \\
  Beijing 100084, China \\
  \texttt{zrt25@mails.tsinghua.edu.cn}
  \And
  Ke Deng\thanks{Corresponding author.} \\
  Department of Statistics and Data Science \\
  Tsinghua University \\
  Beijing 100084, China \\
  \texttt{kdeng@tsinghua.edu.cn}
}

\begin{document}

\maketitle

\begin{abstract}

Density estimation in high-dimensional settings is an important and challenging statistical problem.
Traditional methods based on kernel smoothing are inefficient in high dimensions due to the difficulties in specifying appropriate location-adaptive kernels. 
In this work, we introduce pre-training, a key idea behind many cutting-edge AI technologies, to the context of non-parametric density estimation. 
By establishing a pre-trained neural network that can recommend an appropriate location-adaptive kernel for each sample point, efficient density estimation with adaptive kernels is achieved in high dimensions. 
A wide range of numerical experiments show that this strategy is highly effective for improving density-estimation accuracy, when the target distribution is close to the distribution family for pre-training. 
When the target distribution is substantially different from the pre-training distribution family, the benefit from the proposed pre-training strategy may be diluted, but can be reactivated by an additional fine-tuning procedure.
\end{abstract}


\section{Introduction}
\label{sec:intro}

Nonparametric density estimation is a fundamental problem in statistics and machine learning. 
Given $n$ i.i.d. samples $X_{1:n}=(X_1,\cdots,X_n)$ from a continuous target distribution $F_0$ on $\mathbb{R}^d$, the goal is to estimate $f_0$, the density function of $F_0$, in a non-parametric fashion.
In the statistical literature, a rich collection of methods has been proposed for this important problem, including \emph{kernel density estimation} (KDE) \citep{rosenblatt1956remarks,parzen1962estimation} and smoothing splines \citep{wahba1990spline,green1994nonparametric}.
In parallel, modern machine learning has developed expressive neural density estimators, including autoregressive models \citep{germain2015made,papamakarios2017masked}, normalizing flows \citep{dinh2016density}, and score-based generative models \citep{song2021score}. 
Among these methods, KDE
remains attractive in many applications, because it enjoys an explicit analytic form with easy evaluation and well-developed statistical theory \citep{silverman1986density,wand1994kernel}. 

Like many other methods in this field, however, KDE typically suffers significant performance degradation due to the curse of dimensionality.
When $d$, the dimensionality of the target distribution, is large, the geometry of $f_0$ is difficult to capture, and the kernel function 
controls not only the amount of smoothing, but also the orientation and anisotropy of the estimator \citep{wand1994kernel,chacon2018multivariate}.
In these cases,  using a common global kernel to smooth all regions of the sample space is often inadequate, especially when the target density has heterogeneous local scale or anisotropic geometry: sharp high-density regions may require narrow and directionally sensitive kernels, while sparse or diffuse regions may require flatter kernel functions. 

Adaptive KDE methods \citep{terrell1992variable} mitigate this problem by allowing the bandwidth to vary across locations, with two common formulations being balloon estimators and sample-point estimators \citep{jones1990variable}.
Existing adaptive KDE methods often construct these adaptive bandwidths from pilot density estimates \citep{abramson1982bandwidth}, nearest-neighbor distances \citep{breiman1977variable}, or other local geometric information.
Although statistically well motivated, these approaches typically adapt restricted kernel structures rather than learning a general rule to recommend appropriate location-adaptive kernels.


In this work, we fill this gap by introducing {\bf pre-training}, a key idea behind many cutting-edge AI technologies, to the context of non-parametric density estimation. 
By establishing a pre-trained neural network $\phi_\theta$ with $\theta$ as parameters that can recommend an appropriate location-adaptive kernel 
for each observed sample $X_i$ based on its local geometry in $X_{1:n}$,
we successfully resolve the problem of location-adaptive kernel specification that is often extremely difficult in the traditional learning framework with a limited number of samples. 
With the support of $\phi_\theta$, which encodes information learned during pre-training beyond the limited information contained in the observed sample $X_{1:n}$, efficient adaptive kernel density estimation can be achieved in relatively high-dimensional settings.
To establish the pre-trained neural network $\phi_\theta$ for kernel recommendation, we prepare a distribution family $\cF$ that covers a wide range of distributions of potential interest, and generate a large collection of synthetic data $\cD$
based on $\cF$ for pre-training of $\phi_\theta$.

Numerical experiments show that this \emph{neural-network-guided KDE} (NNKDE) significantly outperforms existing KDE methods when the target distribution $F_0$ is close to $\cF$.
When $F_0$ is far away from $\cF$, the advantage of NNKDE may be diluted, but it can be reactivated by an additional {\bf fine-tuning} procedure, where the global scale of the recommended location-adaptive kernels can be further adjusted under the guidance of self-exclusive leave-one-out likelihood of $X_{1:n}$.
Such a ``pre-training + fine-tuning'' strategy further enhances the adaptability of the proposed NNKDE, providing a powerful tool for non-parametric density estimation.


Our contributions are threefold. 
First, we formally introduce the idea of pre-training to non-parametric statistics and demonstrate its ability to improve the efficiency of statistical inference using synthetic data.
Second, we show that the benefits of pre-training for non-parametric statistics can be further preserved and enhanced by an additional fine-tuning procedure, suggesting that the widely used ``pre-training + fine-tuning'' strategy may also be beneficial for statistical models.
Third, we provide a systematic evaluation protocol to empirically compare different KDE methods in various settings.



\section{Related Work}
\label{sec:related}

\textbf{Bandwidth selection and adaptive KDE.}
Kernel density estimation has a long history in nonparametric statistics, with classical treatments covering both theoretical properties and practical bandwidth selection \citep{silverman1986density,wand1994kernel}. In multivariate KDE, the bandwidth matrix is the central smoothing parameter, controlling scale, orientation, and anisotropy of the estimator \citep{wand1994kernel,chacon2018multivariate}. Classical bandwidth-selection rules include rule-of-thumb methods, plug-in selectors, and cross-validation-based criteria \citep{silverman1986density,sheather1991reliable,rudemo1982empirical,bowman1984alternative}. Extensions to full bandwidth matrices have also been studied, including data-driven smooth cross-validation methods \citep{duong2005cross}. However, a single global bandwidth matrix can be inadequate for heterogeneous densities with spatially varying smoothness, and bandwidth optimization can be numerically unstable in multivariate settings \citep{hall1991local}. Adaptive KDE methods address this limitation by allowing the bandwidth to vary locally. Classical examples include balloon estimators \citep{loftsgaarden1965nonparametric} and sample-point estimators \citep{breiman1977variable,sain2002multivariate}. The latter are particularly relevant here because, under standard constructions, they retain a properly normalized density form \citep{terrell1992variable}. Existing adaptive methods often rely on pilot estimates or scalar local rescaling rules, such as Abramson's square-root law \citep{abramson1982bandwidth,terrell1992variable}. In contrast, our work learns observation-specific positive definite bandwidth matrices within an explicit sample-point KDE.

\textbf{Scalable smoothing splines and adaptive basis selection.}
Smoothing splines and smoothing spline ANOVA models form another important class of classical nonparametric smoothers, but full-basis fitting and smoothing-parameter selection can be computationally demanding in large samples. 
Recent work has addressed this computational bottleneck through adaptive and geometry-aware basis reduction, including adaptive basis sampling for smoothing splines \cite{ma2015efficient}, adaptive basis selection for exponential-family smoothing splines \cite{ma2017adaptive}, space-filling basis selection \cite{meng2020more}, and Hilbert-curve basis selection \cite{meng2022hilbert}. Complementary work has developed scalable smoothing-parameter selection procedures for smoothing spline ANOVA models \cite{helwig2015fast, sun2021asymptotic}. These studies are closely related to our work in their shared goal of making classical nonparametric estimators more scalable and adaptive. However, they mainly approximate penalized-spline regression or ANOVA estimators, or accelerate their smoothing-parameter selection. In contrast, NNKDE learns a transferable rule for assigning observation-specific positive definite bandwidth matrices within an explicit sample-point KDE.

\textbf{Neural density estimation.}
A separate line of work uses neural networks to construct expressive density estimators. Autoregressive density estimators such as Masked Autoencoder for Distribution Estimation (MADE) model the joint density through conditional factorizations \citep{germain2015made}. Normalizing flows, including Real-valued Non-Volume Preserving transformations (Real NVP) \citep{dinh2016density} and masked autoregressive flow (MAF) \citep{papamakarios2017masked}, learn invertible transformations from simple base distributions to complex target distributions. Score-based generative models learn the score function and generate samples through stochastic dynamics \citep{song2021score}. Other neural approaches have been developed for high-dimensional density estimation using deep generative models \citep{liu2021density}. These approaches rely on neural approximation, but they primarily target direct density, score, or generative-map estimation. NNKDE differs in that the neural component is embedded inside an explicit KDE estimator rather than replacing the estimator itself.

\textbf{Amortized and pre-trained density learning.}
The pre-training strategy in this paper is related to amortized inference, where computation is shared across a family of related inference problems rather than repeated from scratch for each new instance \citep{gershman2014amortized}. A similar perspective appears in simulation-based inference, where simulation and surrogate learning are performed offline and then reused for new observations \citep{cranmer2020frontier}. Related ideas have also been explored in density estimation, including data-driven density estimators trained over synthetic distribution families \citep{puchert2021datadriven}, meta-learned conditional density estimators across related tasks \citep{ton2021noise}, and transformer-based plug-in estimators for density and score estimation \citep{ilin2025discoformer}. These approaches amortize density or score estimation directly. In contrast, NNKDE amortizes the bandwidth-selection rule inside a classical sample-point KDE, so pre-training is used to learn a transferable bandwidth functional rather than to replace the nonparametric estimator.

\section{Method}
\label{sec:method}

\begin{figure}[t]
\centering
\includegraphics[width=\textwidth]{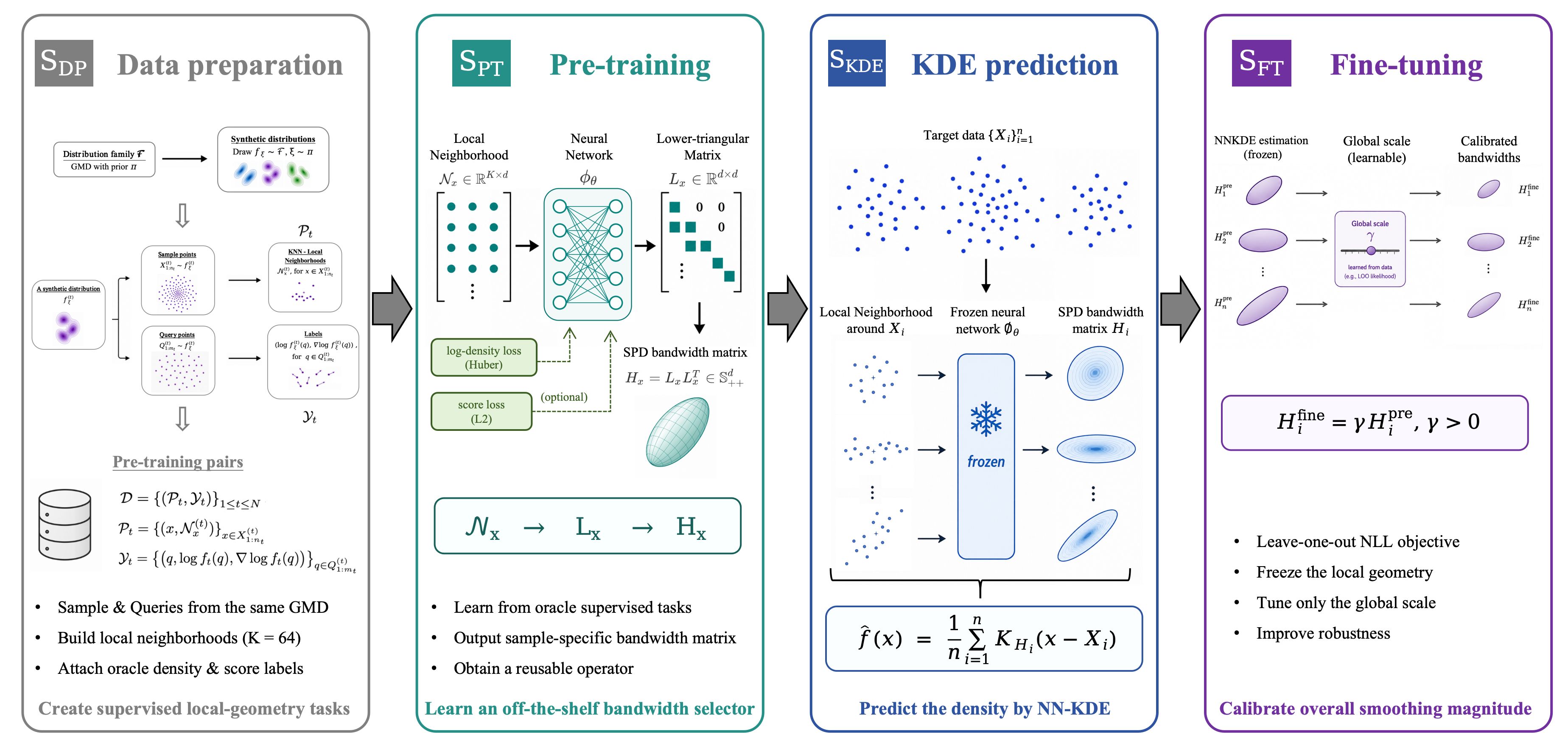}
\caption{Workflow of NNKDE composed of four stages. 
}
\label{fig:method_overview}
\end{figure}

In this work, we consider the following sample-point adaptive KDE for the target distribution $F_0$ based on i.i.d. samples $X_{1:n}\sim F_0$:
\begin{equation}\label{eq:adaptiveKDE}
\widehat f_{\text{SP}}(x\mid X_{1:n})=\frac{1}{n}\sum_{i=1}^n K_{H_i}(x-X_i)\qquad \forall\ x\in\bbR^d,
\end{equation}
where $H_i$ is a $d\times d$ covariance matrix (referred to as the bandwidth matrix hereinafter) associated with the sample point $X_i$, and $K_H(z)=|H|^{-1/2}\kappa(H^{-1/2}z)$ with $\kappa$ being the $d$-dimensional standard Gaussian kernel.
The bandwidth matrix $H_i$ fully controls the geometric shape of the location-adaptive kernel associated with $X_i$.
The proposed NNKDE relies on a pre-trained neural network $\phi_\theta$ to recommend $H_i$ for $X_i$.
Figure~\ref{fig:method_overview} demonstrates the workflow of NNKDE, which is composed of four stages: the data preparation stage $\cS_\text{DP}$, the pre-training stage $\cS_\text{PT}$, the KDE stage $\cS_\text{KDE}$, and the fine-tuning stage $\cS_\text{FT}$.
Next, we will introduce these four stages in sequence.


\subsection{The pre-training data preparation stage $\cS_\text{DP}$}

This stage concerns preparation of the distribution family $\cF$ and the synthetic data $\cD$ defined in the Introduction for pre-training of $\phi_\theta$.
Here, we specify $\cF$ as a family of \emph{Gaussian mixture distributions} (GMDs), i.e., $\cF=\{\text{GMD}_\xi\}_{\xi\sim\pi}$, due to their ease of operation and flexibility in distribution approximation, where a GMD parameter $\xi$ is randomly generated from a prior distribution $\pi$ over the parameter space $\Xi$.
In practice, we randomly generate $N$ GMDs from $\cF$, namely $\{F_t\}_{1\leq t\leq N}$, where $F_t=\text{GMD}_{\xi_t}$ with $\xi_t\sim\pi$.
Concrete specifications of $\Xi$ and $\pi$ are detailed in Appendix~\ref{app:data_pretraining}, which ensures the rich diversity of $\cF$.

For each obtained $F_t$, we draw a sample set $X_{1:n_t}^{(t)}\sim F_t$ 
together with an independent query set $Q_{1:m_t}^{(t)}\sim F_t$, 
leading to the following triplets $\{(F_t, X_{1:n_t}^{(t)}, Q_{1:m_t}^{(t)})\}_{1\leq t\leq N}$.
For each sample point $x\in X_{1:n_t}^{(t)}$, let $\cN_x^{(t)}$ denote the $K\times d$ coordinate matrix of the $K$ nearest neighbors of $x$ in $X_{1:n_t}^{(t)}$ after a translation transformation of the sample space that moves $x$ to the origin. 
The matrix $\cN_x^{(t)}$ encodes the local geometric pattern around $x$ in contrast to the background $X_{1:n_t}^{(t)}$,
and 
\begin{equation}\label{eq:cP_i}
\cP_t=\{(x,\cN_x^{(t)})\}_{x\in X_{1:n_t}^{(t)}}
\end{equation}
summarizes all local patterns of $X_{1:n_t}^{(t)}$.
For each query point $q\in Q_{1:m_t}^{(t)}$, let $\log f_t(q)$ and $\nabla\log f_t(q)$ be logarithmic density and score of $F_t$ at $q$, respectively.
The collection of triplets
\begin{equation}
\cY_t=\big\{\big(q,\log f_t(q),\nabla\log f_t(q)\big)\big\}_{q\in Q_{1:m_t}^{(t)}}
\label{eq:oracle_query_labels}
\end{equation}
summarizes all order-0 and order-1 information about $f_t$, the density function of $F_t$, at $Q_{1:m_t}^{(t)}$.
Pairing $\cP_t$ and $\cY_t$ for each $t$, we obtain the following data set for the pre-training of $\phi_\theta$:
\begin{equation}
\cD=\left\{(\cP_t, \cY_t)\right\}_{1\leq t\leq N},
\label{eq:pretraining_task_collection}
\end{equation}
where $\cY_t$ serves as the training guidance for input $\cP_t$.
In this study, we generated a large-scale pre-training data set via these steps by specifying $N=500,000$, and $n_t=2048$ and $m_t=1024$ for all $t$.


\subsection{The pre-training stage $\cS_\text{PT}$}
This stage concerns learning the neural kernel recommender $\phi_\theta$, whose architecture is detailed in Appendix~\ref{app:architecture}, from the pre-training data set $\cD$ prepared in the previous stage. 
For each element $(\cP_t, \cY_t)$ in $\cD$, neural network $\phi_\theta$ digests $\cN_x^{(t)}$ and recommends a lower-triangular matrix $L_x$ determined by $\theta$ for each $x\in X^{(t)}_{1:n_t}$, which leads to a bandwidth matrix $H_x=L_xL_x^T$.
We denote this procedure by the notation $H_x=\phi_\theta(\cN_x)$, with the note that $H_x$ is a function of $\theta$.
According to \eqref{eq:adaptiveKDE}, these location-adaptive bandwidth matrices induce the following estimator for the fully known $f_t$:
\begin{equation}
\widehat f_\theta(x\mid X_{1:n_t}^{(t)})=
\frac{1}{n_t}\sum_{l=1}^{n_t}K_{H_{l}^{(t)}}(x-X_l^{(t)})
\quad\text{with}\quad
H_{l}^{(t)}=\phi_\theta(\cN_{X_l^{(t)}}^{(t)}),
\label{eq:pretrain_induced_kde}
\end{equation}
whose score function has the closed form below:

$$\nabla_x \log \widehat f_{\theta}(x\mid X_{1:n_t}^{(t)})=\sum_{l=1}^{n_t} \omega_l(x)\left\{[H_l^{(t)}]^{-1}(X_l^{(t)}-x)\right\},\quad 
\omega_l(x)=\frac{K_{H_l^{(t)}}(x-X_l^{(t)})}{\sum_{r=1}^{n_t} K_{H_r^{(t)}}(x-X^{(t)}_r)}.$$

Comparing $\widehat f_\theta$ with $f_t$ at query points $Q^{(t)}_{1:m_t}$, we obtain the following loss function w.r.t. $(\cP_t, \cY_t)$:
\begin{equation}
\begin{aligned}
\cL_t(\theta)&=\frac{1}{m_t}\sum_{j=1}^{m_t} \rho\!\left(\log \widehat f_\theta(Q_j^{(t)}\mid X_{1:n_t}^{(t)})-\log f_t(Q_j^{(t)})\right) \\
&\quad+\lambda_{\mathrm{score}}\cdot\frac{1}{m_t}\sum_{j=1}^{m_t}\left\|\nabla \log \widehat f_\theta(Q_j^{(t)}\mid X_{1:n_t}^{(t)})-\nabla \log f_t(Q_j^{(t)})\right\|_2^2,
\end{aligned}
\label{eq:pretrain_loss}
\end{equation}
where $\rho$ is the Huber loss, $\|\cdot\|_2$ is the $L_2$ loss, and $\lambda_{\mathrm{score}}\geq 0$ controls the strength of the second term.
Intuitively, the first term in $\cL_t(\theta)$ guides $\phi_\theta$ to produce bandwidth matrices that yield accurate density values, while the score term provides additional supervision on the higher-order geometry of $f_t$.
To keep a proper balance between these two terms, we specify $\lambda_{\mathrm{score}}=10^{-2}$ in this study.

Assembling the hybrid loss from different elements in $\cD$, we obtain the overall loss to guide the pre-training of $\phi_\theta$, based on which $\theta$ can be optimized by AdamW  \citep{loshchilov2018decoupled} under the default setting with learning rate and weight decay both specified as $10^{-4}$.
In practice, however, to avoid memory overflow, we follow the mini-batch strategy to update $\theta$ according to the assembled loss of $B=500$ elements in $\cD$ each epoch.
It takes 1,000 epochs to scan through the pre-training set $\cD$ with $N=500,000$ elements. 

For each dimension $d$, we pre-trained a separate model on a single NVIDIA Tesla V100-SXM2-32GB GPU. Pre-training took approximately 15 GPU-hours for each dimension up to $d=10$, and approximately 19, 27, and 39 GPU-hours for $d=20$, $d=30$, and $d=50$, respectively.



\subsection{The KDE stage $\cS_\text{KDE}$}
\label{subsec:learned_bandwidths}

Let $\widehat\theta$ be the optimized parameter obtained in the pre-training stage.
Neural network $\phi_{\widehat\theta}$ offers an off-the-shelf bandwidth-selection rule in downstream applications. 
Given the target sample $X_{1:n}\sim F_0$, $\phi_{\widehat\theta}$ recommends for each $X_i$ in $X_{1:n}$ the following bandwidth matrix
\begin{equation}
H_i^{\mathrm{pre}}=\phi_{\widehat\theta}(\cN_{X_i}), \qquad i=1,\cdots,n,
\label{eq:target_pretrained_bandwidth}
\end{equation}
where $\cN_x$ denotes the $K\times d$ coordinate matrix of the $K$ nearest neighbors of $x$ in $X_{1:n}$ after a translation transformation of the sample space that moves $x$ to the origin. 
These bandwidth matrices lead to the following estimator for $f_0$:
\begin{equation}
\widehat f_{\mathrm{pre}}(x\mid X_{1:n})=\frac{1}{n}\sum_{i=1}^n K_{H_i^{\mathrm{pre}}}(x-X_i),\qquad \forall\ x\in\bbR^d.
\label{eq:pretrained_kde}
\end{equation}
No target-specific retraining of the neural network is performed in this stage.




\subsection{The fine-tuning stage $\cS_\text{FT}$}
\label{subsec:pretraining_adaptation}

When the target distribution differs from the pre-training family, the global smoothing scale of the pre-trained bandwidth matrices may need adjustment. 
We therefore adopt a lightweight scale-calibration step as the ``fine-tuning'' for the pre-trained $\phi_{\widehat\theta}$. 
For this purpose, we modify the bandwidth matrices recommended by $\phi_{\widehat\theta}$ with a global scalar $\gamma>0$ as below:
\begin{equation}
H_i^{\mathrm{fine}}=\gamma H_i^{\mathrm{pre}}, \qquad i=1,\cdots,n.
\label{eq:scale_adaptation}
\end{equation}

The optimal $\gamma$ is obtained by minimizing a self-exclusive leave-one-out negative log-likelihood on the target sample, i.e.,
\begin{equation}
\gamma^\star=\arg\min_{\gamma>0}\left\{-\frac{1}{n}\sum_{i=1}^n
\log\left[\frac{1}{n-1}\sum_{j\neq i}K_{\gamma H_j^{\mathrm{pre}}}(X_i-X_j)\right]\right\}.
\label{eq:scale_adapt_objective}
\end{equation}
The self-exclusion form removes the direct contribution of the kernel centered at $X_i$, protecting $\gamma$ from being driven toward degenerate undersmoothing.
The optimization problem defined in \eqref{eq:scale_adapt_objective} can be solved conveniently by gradient-based optimization methods.
Plugging $\gamma^\star$ into \eqref{eq:scale_adaptation}, we obtain the fine-tuned bandwidth matrices to improve adaptive KDE.
This fine-tuning step keeps the pre-trained neural bandwidth recommender fixed and only calibrates a single global bandwidth scale on the target sample.
Hereinafter, we refer to NNKDE without fine-tuning as NNKDE$_{\mathrm{pre}}$ and NNKDE with global scale calibration as NNKDE$_{\mathrm{fine}}$.

\section{Experiments}
\label{sec:experiments}
In this section, we compare the performance of the proposed methods, i.e., NNKDE$_{\mathrm{pre}}$ and NNKDE$_{\mathrm{fine}}$, with a collection of representative KDE and adaptive KDE methods in the literature.
For KDE methods, we select the widely used Silverman's rule \citep{silverman1986density} and likelihood cross-validation (LCV) \citep{rudemo1982empirical,bowman1984alternative} as comparison baselines;
for adaptive KDE methods, we select Abramson's method \citep{abramson1982bandwidth,terrell1992variable} and kNN local-scaling KDE (kNN) \citep{loftsgaarden1965nonparametric,breiman1977variable} as comparison baselines.
A popular neural density estimator MAF \citep{papamakarios2017masked} is compared as well.
Moreover, to evaluate the effect of pre-training in NNKDE, we also include NNKDE$_{\mathrm{scratch}}$, a naive version of NNKDE where the parameter $\theta$ of the neural network $\phi_\theta$ is randomly initialized without further training, in the comparison. 
More details on the implementation of these baseline methods are provided in Appendix~\ref{app:baselines}.

\subsection{Experimental setup}
\label{subsec:exp_setup}

\begin{figure*}[hpbt]
\centering
\includegraphics[width=\textwidth]{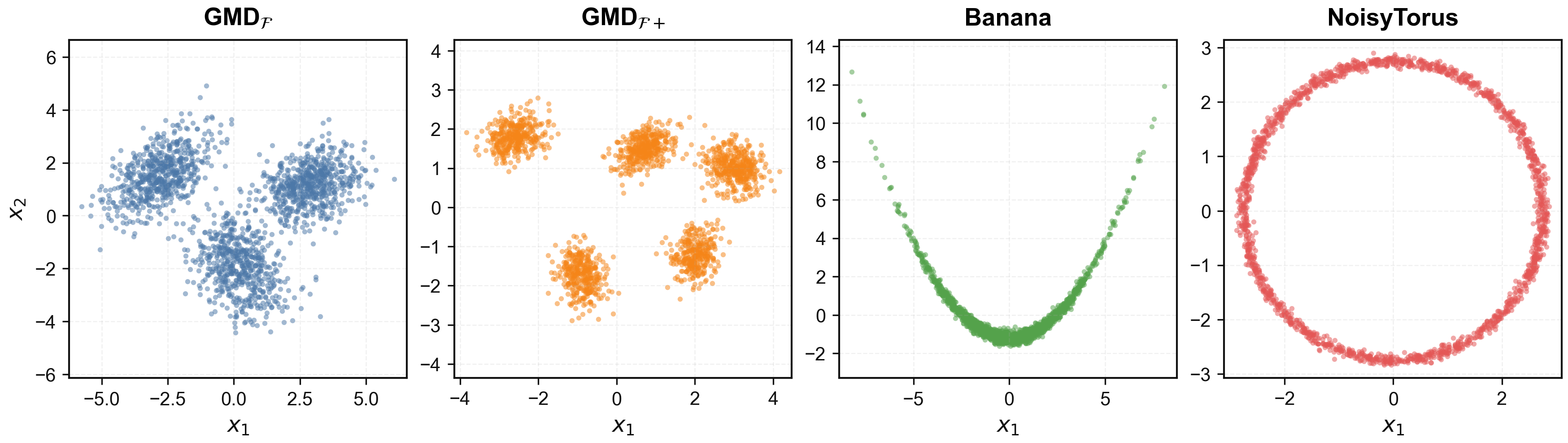}
\caption{Representative two-dimensional samples from the four benchmark scenarios.}
\label{fig:typical_scenarios_d2}
\end{figure*}

To systematically compare the performance of different methods, we set up 4 scenarios in which the target distribution $F_0$ comes from 4 different distribution families, namely GMD$_{\cF}$, GMD$_{\cF+}$, Banana and NoisyTorus. Figure \ref{fig:typical_scenarios_d2} presents the typical scatter plots of the four scenarios when $d=2$.
In the first scenario GMD$_{\cF}$, the target distribution $F_0$ is generated from $\cF=\{\text{GMD}_\xi\}_{\xi\sim\pi}$, the distribution family used in the pre-training of $\phi_\theta$.
In all the other 3 scenarios, the target distribution $F_0$ is generated from alternative distribution families that do not overlap with $\cF$.
More details about these scenarios are provided in Appendix~\ref{app:scenarios}.

For each scenario, we choose the dimensionality $d\in\{2,3,5,10,30,50\}$, and specify the sample size $n \in\{64, 256, 1024, 2048, 4096\}$ for low dimensional cases where $d\leq 5$, while $n \in\{2000, 5000, 10000, 15000, 30000\}$ for high dimensional cases where $d>5$, resulting in $4\times (3\times 5+3\times 5)=120$ experimental settings.


For each experimental setting, we independently generate 10 target distribution instances. 
Within each target distribution instance, we draw 5 independent fitting samples and evaluate the resulting estimators on independent test sets sampled from the same distribution.
This yields 50 runs for each experimental setting in total.
We apply the 8 competing methods, namely Silverman, LCV, Abramson, kNN, MAF, NNKDE$_{\text{scratch}}$, NNKDE$_{\text{pre}}$, NNKDE$_{\text{fine}}$, to all runs, and evaluate their performance by the normalized out-of-sample negative log-likelihood (NLL) defined below:
\begin{equation}
\mathrm{NLL}(\widehat f)=-\frac{1}{N_{\mathrm{eva}}}\sum_{j=1}^{N_{\mathrm{eva}}}\log \widehat f(Z_j)/d,
\label{eq:app_nll_empirical}
\end{equation}
where $\widehat f$ denotes the density estimator obtained from a specific method, and $\{Z_j\}_{j=1}^{N_{\mathrm{eva}}}$ is an independent test set sampled from the corresponding target distribution instance. 
We set $N_{\mathrm{eva}}=3000$ in all experiments. 
A smaller NLL indicates better out-of-sample likelihood performance, and normalization by the ambient dimension $d$ facilitates comparison across dimensions. 
For each experimental setting, we report the mean and standard deviation across the 10 target distribution instances after averaging over the 5 replicates within each instance.

Moreover, to better judge how the competing methods approach the theoretical optimal performance, we also evaluate the performance of the true density $f_0$ under the same evaluation protocol.
We refer to the true-density evaluation as the Oracle reference, which helps highlight the performance gap between NNKDE and theoretical optimum.

\subsection{Results}
\label{subsec:Results}

\begin{figure*}[htbp]
\centering
\includegraphics[width=\textwidth]{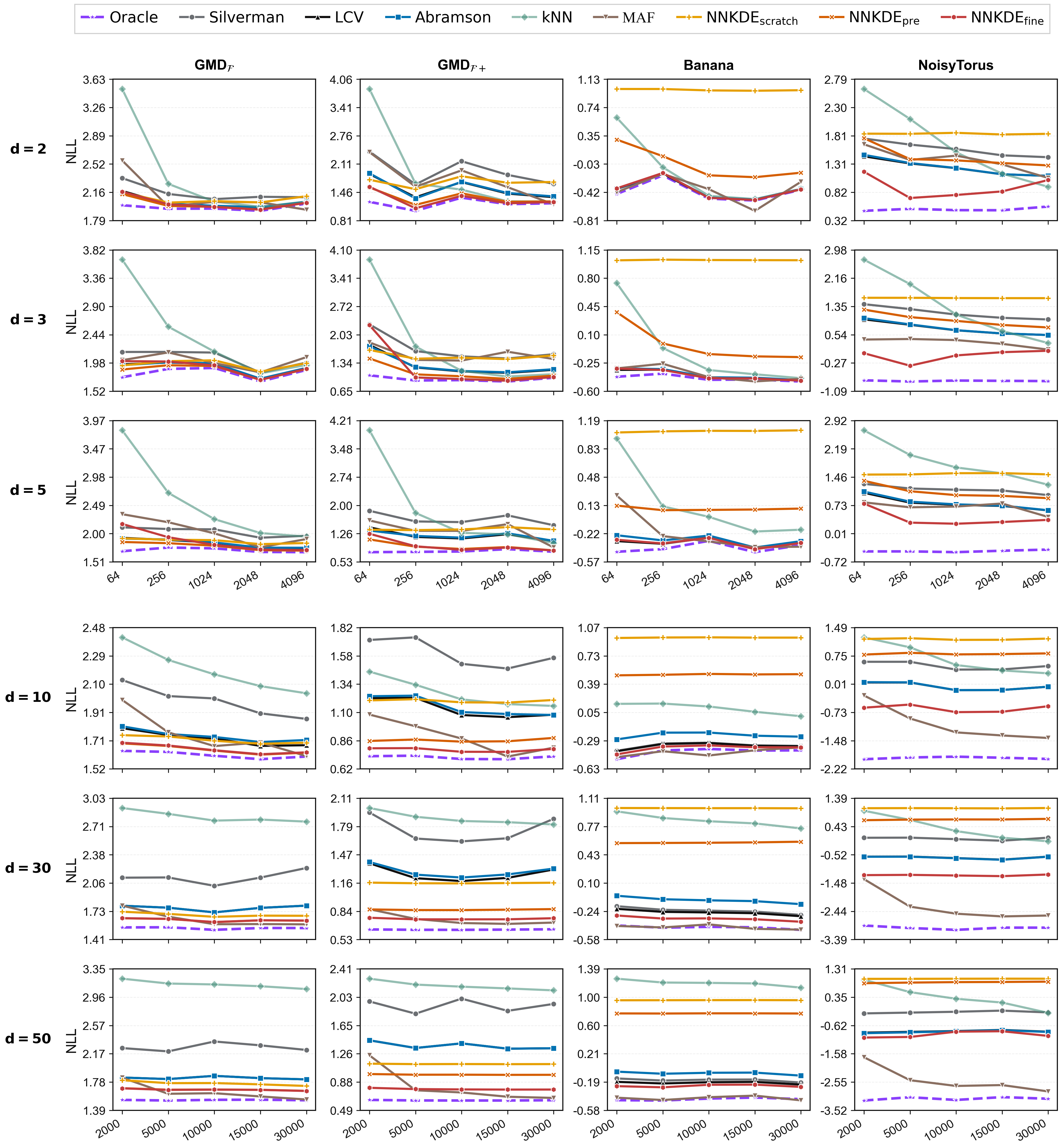}
\caption{Performance of different methods in terms of normalized negative log-likelihood.}
\label{fig:ss_free_y}
\end{figure*}

Figure~\ref{fig:ss_free_y} compares the performance of different KDE methods in terms of average NLL in four evaluation scenarios with different combinations of $d$ and $n$.
From the figure, we observe the following patterns.
First, NNKDE$_{\text{fine}}$ generally outperforms the competing KDE methods across most settings in almost all experimental settings, indicating that the proposed ``pre-training + fine-tuning'' framework is a promising solution to the challenging problem of adaptive KDE.
Second, NNKDE$_{\text{fine}}$ yields competitive performance with respect to Oracle and MAF in almost all cases of scenarios GMD$_{\cF}$ and $GMD_{\cF+}$ and most low-dimensional cases of scenarios Banana and NoisyTorus, but maintains a distance to Oracle and MAF in high-dimensional cases of scenarios Banana and NoisyTorus, suggesting that insufficient pre-training may have a negative impact on the performance of NNKDE in practice.
Third, NNKDE$_{\text{pre}}$ yields competitive performance in scenarios GMD$_\cF$ and GMD$_{\cF+}$ (almost identical to NNKDE$_{\text{fine}}$ in scenarios GMD$_\cF$ as expected) and obvious performance degradation in scenarios Banana and NoisyTorus, confirming that the fine-tuning stage is critical to the success of NNKDE$_{\text{fine}}$.
Fourth, NNKDE$_{\text{scratch}}$ performs poorly in most cases, revealing the important role of the pre-training stage in the proposed methods.

More detailed results are reported in Appendix~\ref{app:extra_dim_scaling}.
Table~\ref{tab:ds_nll_per_dim_all_dim} provides additional comparison of different KDE methods in the cross-section of $n=4,096$ with more numerical details.
Each cell in the table reports two numbers for an experimental setting: the average NLL and the corresponding standard deviation.
In general, the table delivers similar messages as in Figure~\ref{fig:ss_free_y}.
A notable observation from this table is that the NNKDE methods tend to exhibit smaller performance variation than the other KDE methods in most high-dimensional cases. 
Moreover, Figure~\ref{fig:fit_time_ss_log10} compares the running time of different methods in seconds on the $\log_{10}$ scale.
The fitting time of the NNKDE methods covers the KDE prediction stage and the fine-tuning stage, without considering the pre-training stage.
From Figure~\ref{fig:fit_time_ss_log10} we can see that the proposed NNKDE methods are substantially faster than MAF, because they do not rely on freshly prepared neural networks as in MAF.
Together with the results in Figure~\ref{fig:ss_free_y}, these findings demonstrate the potential of NNKDE as a framework that integrates neural networks with statistical inference through pre-training and fine-tuning.

\section{Conclusions and Discussions}
\label{sec:disc_conc}

We proposed NNKDE, a sample-point adaptive KDE method that adopts reusable kernel selection rules encoded in a pre-trained neural network, while preserving the explicit KDE form. 
Systematic numerical experiments confirm the effectiveness of the proposed method in improving the accuracy of non-parametric density estimation, and the critical role of the pre-training and fine-tuning stages built in the approach.
These results suggest a promising way to bring amortized pre-training into adaptive nonparametric density estimation and beyond, which may reshape many areas of non-parametric statistics. 

Although the pre-training stage of NNKDE is computationally very expensive, once the pre-training is finished, applying NNKDE in practice is computationally convenient.
This paradigm of heavy offline computing but light online computing keeps a perfect balance between pursuit of excellent performance and saving data and computing resources.
This property grants NNKDE great advantages in practical applications with a limited computation budget.

Unlike other neural KDE methods, which directly approximate the unknown density function via a neural network, NNKDE utilizes a neural network to learn a more fundamental bandwidth selection rule from local geometric properties of the distribution family for pre-training in advance, which can be further adjusted via fine-tuning after the real task is received.
Such a pipeline is an exact analogy to the ``pre-training + fine-tuning'' framework which has been widely adopted in image and natural language processing.
On the other hand, because NNKDE takes KDE as the primary framework to integrate a pre-trained neural network, it retains the transparency and interpretability typically associated with statistical models but often absent from deep learning models.

We believe that the proposed NNKDE method has a similar convergence order as classic adaptive KDE methods, such as the Abramson-type estimators, in the general cases where the target distribution is intrinsically $d$-dimensional, but enjoys a higher convergence rate because of more efficient bandwidth matrix selection due to pre-training. 
In the degenerate case where the target distribution lives in a lower-dimensional manifold, NNKDE may enjoy a higher convergence order than classic adaptive KDE methods because it is more capable to discover the intrinsic dimensionality of the target distribution.

Despite the many advantages mentioned above, the NNKDE introduced in this paper still has many limitations.
First, the pre-training dataset is limited to Gaussian mixtures with only a few Gaussian components, greatly restricting the pre-training module from releasing greater power.
Second, the current architecture of the neural kernel recommender $\phi_\theta$ is not flexible enough to support a varying ambient dimension $d$.
Such a limitation makes it inefficient to utilize cross-dimensional pre-training information and deal with degenerate distributions, e.g., those in the NoisyTorus scenario, whose probability mass is concentrated near a lower-dimensional manifold of the space.
Third, the current fine-tuning strategy to adjust a single global scalar factor is too naive to meet the heterogeneous fine-tuning demand in different local regions of the target family.
More flexible fine-tuning mechanisms need to be investigated to further improve effectiveness and adaptability of pre-training.

NNKDE utilizes synthetic data heavily in its pre-training stage.
However, unlike most recent work that constructs synthetic data for model training, NNKDE uses synthetic data for pre-training instead.
This new paradigm of using synthetic data introduces many new perspectives and problems regarding the construction and utilization of synthetic data.
When synthetic data are used for model training, they are often interpreted as an informal prior from a Bayesian perspective, and their impact on statistical inference is studied under distributional drift between real and synthetic data.
When synthetic data are used for pre-training followed by fine-tuning, a new framework is needed because the role of synthetic data becomes more subtle.

As a new effort to integrate deep learning and statistical learning for more efficient data analysis, NNKDE is unique in its initiative to enhance a complex neural network with pre-training.
Embedding a complex neural network without sufficient pre-training into a statistical model is often dangerous in practice because the limited sample size in a typical statistical problem usually makes it unaffordable to drive the expensive neural network.
In these cases, establishing a wise framework to effectively utilize the rich information in a well pre-trained neural network indirectly, just as what NNKDE achieves, is often an attractive choice.

\section{Acknowledgments}
The authors thank colleagues and seminar participants for helpful discussions and constructive feedback. Any remaining errors are our own.

% ...

\bibliographystyle{plainnat}
\bibliography{staix_2026_sample}

\newpage
\appendix
\setcounter{figure}{0}
\renewcommand{\thefigure}{A\arabic{figure}}
\renewcommand{\figurename}{Fig.}

\setcounter{equation}{0}
\renewcommand{\theequation}{A\arabic{equation}}

\setcounter{table}{0}
\renewcommand{\thetable}{A\arabic{table}}
\renewcommand{\tablename}{Table}

\section{Details on Synthetic Pre-training Data Generation}
\label{app:data_pretraining}
Here, we provide concrete specification of the ``prior'' distribution $\pi$ over $\Xi$, which defines $\cF=\{\text{GMD}_\xi\}_{\xi\sim\pi}$, the family of Gaussian mixture distributions for the pre-training of $\phi_\theta$.
For a GMD $F_\xi$ from $\cF$, its density is
\begin{equation}
f_\xi(x)=\sum_{k=1}^K w_k \mathcal N(x;\mu_k,\Sigma_k),
\label{eq:app_gmm}
\end{equation}
where $\xi=\{K,w_{1:K},\mu_{1:K},\Sigma_{1:K}\}$. 
We specify $\pi$ as follows: the number of mixture components $K\sim\text{Unif}\{1,\dots,8\}$, the weight vectors  $w_{1:K}\sim\mathrm{Dirichlet}(0.8\times {\mathbf 1}_K)$, the component means $\mu_k\sim\mathrm{Unif}([-10,10]^d)$, and the component covariance matrix $\Sigma_k$ is generated by randomly sampling its eigenvalues from $\mathrm{Unif}([0.25,2.5])$ and specifying its eigenvectors with a random rotation matrix in $\bbR^d$ when $d\leq 5$. For $d>5$, diagonal covariance matrices with entries sampled from the same variance range are used for numerical stability.

\section{Architecture of the Neural Kernel Recommender $\phi_\theta$}
\label{app:architecture}

To describe the architecture independently of its use inside KDE, we use $\eta\in\mathbb R^d$ to denote a generic reference location. Let
\begin{equation}
\mathcal N_\eta=\{r_\ell=u_\ell-\eta:u_\ell\in \mathrm{kNN}(\eta)\}_{\ell=1}^{K_{\mathrm{nn}}}\in\mathbb R^{K_{\mathrm{nn}}\times d}.
\label{eq:app_neighbourhood}
\end{equation}
The neural bandwidth recommender maps this local neighbourhood geometry to a symmetric positive definite bandwidth matrix,
\begin{equation}
\phi_\theta:\mathcal N_\eta\mapsto H_\eta .
\label{eq:app_phi_mapping}
\end{equation}
Here $\eta$ is only a generic reference location for describing the architecture. In pre-training, it is instantiated as a point in the fitting sample $X_{1:n_i}^{(i)}$; in the KDE stage, it is instantiated as a target sample point $X_i$.

Internally, the network uses an SPD-constrained output parameterization. Specifically, it outputs a lower-triangular factor $L_\eta$ and sets
\begin{equation}
H_\eta=L_\eta L_\eta^\top .
\label{eq:app_cholesky_param}
\end{equation}
This implementation detail ensures that the bandwidth matrix returned by $\phi_\theta$ is positive definite, while the main text uses the simpler notation $\phi_\theta(\mathcal N_\eta)=H_\eta$.

\begin{figure}[htbp]
\centering
\includegraphics[width=\textwidth]{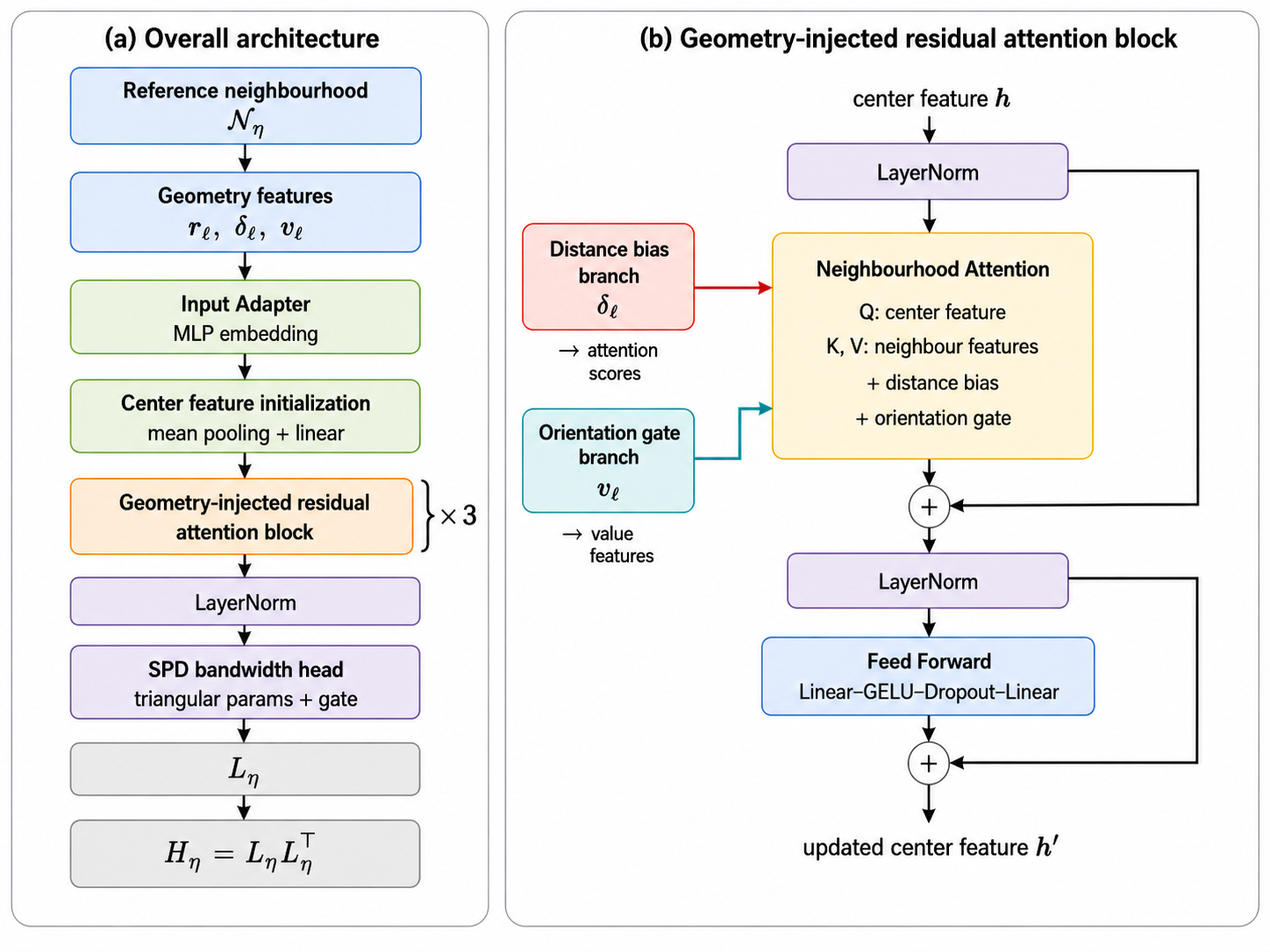}
\caption{Architecture of the neural bandwidth recommender $\phi_{\theta}:\mathcal N_\eta \to H_\eta$ for a generic reference location $\eta\in\mathbb R^d$.}
\label{fig:nn_architecture}
\end{figure}

Each interaction block updates a center feature by attending from the sample-point representation to its neighbour features. The attention scores include a learnable distance bias with a locality prior, and the value features are modulated by an orientation gate. The main experiments use $3$ interaction blocks, $4$ attention heads, hidden dimension $128$, and dropout rate $0.1$.

The bandwidth head outputs $d(d+1)/2+1$ scalars for each sample point. The first $d(d+1)/2$ values parameterize a raw lower-triangular matrix, while the final scalar is passed through a sigmoid gate to control the magnitude of strictly lower-triangular entries. Diagonal entries are clipped and exponentiated to ensure positivity, and strictly lower-triangular entries are scaled by a conservative off-diagonal factor. The resulting lower-triangular factor $L_i$ defines
\begin{equation}
H_i=L_iL_i^\top.
\label{eq:app_spd}
\end{equation}
This Cholesky-type parameterization ensures that each predicted bandwidth matrix is symmetric positive definite. The full-matrix form allows locally anisotropic smoothing, while the SPD constraint ensures that the resulting Gaussian kernels are valid density kernels.

For numerical stability, density evaluation is performed using log-domain summation, and matrix solves are implemented through triangular operations associated with the Cholesky factors rather than explicit matrix inversion. Dimension-dependent numerical defaults are used for diagonal clipping, off-diagonal scaling, and optional shrinkage.

\section{Implementation Details of Baseline Methods}
\label{app:baselines}

This section describes the baseline methods, NNKDE variants, and numerical implementation details used in the experiments.


\paragraph{Scott and Silverman rules (Scott / Silverman).}
Scott's and Silverman's rules are normal-reference bandwidth selectors for KDE \citep{scott1992multivariate,silverman1986density}. They use a single global bandwidth determined from the sample size, dimension, and empirical scale of the data. These methods are included as simple rule-of-thumb KDE baselines.

\paragraph{Global KDE with likelihood cross-validation (LCV).}
The global KDE-LCV baseline uses a single global bandwidth selected by likelihood cross-validation \citep{rudemo1982empirical,bowman1984alternative}. This baseline optimizes the smoothing scale using the observed sample, but it still applies one global bandwidth structure to the entire dataset.

\paragraph{Abramson adaptive KDE (Abramson).}
The Abramson baseline is a classical adaptive KDE method based on local bandwidth rescaling \citep{abramson1982bandwidth,terrell1992variable}. It adjusts the local bandwidth magnitude according to a pilot density estimate. This provides a non-neural sample-point adaptive KDE baseline.

\paragraph{kNN local-scaling KDE (kNN).}
The kNN local-scaling baseline assigns local bandwidth scales using nearest-neighbour distances. It is included as a simple geometry-based adaptive KDE method, related to nearest-neighbour density estimation and sample-point adaptive smoothing \citep{loftsgaarden1965nonparametric,breiman1977variable}.

\section{Scenarios for Method Comparison}
\label{app:scenarios}


\textbf{Scenarios GMD$_\cF$ and GMD$_\cF+$.}
The generation rule of the target distribution $F_0$ in scenario GMD$_\cF$ has been explicitly established in Appendix~\ref{app:data_pretraining}.
In scenario GMD$_\cF+$, we keep the same setting except that the eigenvalue distribution of $\Sigma_k$ is changed from $\mathrm{Unif}([0.25,2.50])$ to $\mathrm{Unif}([0.15,0.25])$.
This modification leads to more compact Gaussian components than in the GMD$_\cF$ scenario.

\textbf{Scenario Banana.} In this scenario, we generate the target distribution $F_0$ from the \emph{Banana-shaped family}, a nonlinear non-Gaussian distribution family parameterized by a curvature parameter $b$ and coordinate-wise noise standard deviations $\sigma_{1:d}$. Specifically, for each sample
$X_i=(X_{i,1},\ldots,X_{i,d})\in\mathbb R^d$, the coordinates are generated autoregressively as follows:
\begin{equation}
X_{i,1}\sim N(0,\sigma_1^2),\qquad 
X_{i,k}\mid X_{i,k-1}\sim 
N\big(b(X_{i,k-1}^2-\sigma_{k-1}^2),\sigma_k^2\big),
\quad k=2,\dots,d.
\label{eq:app_banana}
\end{equation}
This construction creates nonlinear dependence across coordinates through the quadratic drift term while retaining a tractable density by the autoregressive factorization
\begin{equation}
    f_0(x_i)=f_1(x_{i,1})\prod_{k=2}^d f_k(x_{i,k}\mid x_{i,k-1}).
\end{equation}

In the experiment, we sample $b\sim \mathrm{Unif}[0.08,0.35]$ and independently sample
$\sigma_k\sim \mathrm{Unif}[0.05,0.30]$ for $k=1,\dots,d$.

\textbf{Scenario NoisyTorus.}
In this scenario, we generate the target distribution $F_0$ from a \emph{noisy torus-type family}, whose probability mass concentrates near a low-dimensional circular manifold embedded in $\mathbb R^d$. The construction uses one noisy circular block when $d=2$ or $3$, and two independent noisy circular blocks when $d\geq 4$. Let $k_d=1$ for $d<4$ and $k_d=2$ for $d\geq 4$, and let $r_d=d-2k_d$ be the number of remaining Gaussian noise coordinates.

For $j=1,\ldots,k_d$, generate
\begin{equation}
Z_j=(R_j+\sigma_r\rho_j)(\cos\theta_j,\sin\theta_j)^\top, \qquad \theta_j\sim\mathrm{Unif}(0,2\pi), \qquad \rho_j\sim N(0,1).
\end{equation}
If $r_d>0$, generate $E\sim N(0,I_{r_d})$ and set $Y=(Z_1^\top,\ldots,Z_{k_d}^\top,\sigma_\perp E^\top)^\top$; otherwise set $Y=(Z_1^\top,\ldots,Z_{k_d}^\top)^\top$. The final observation is
\begin{equation}
X=\mu+QY,
\end{equation}
where $Q$ is a random orthogonal matrix and $\mu\sim\mathrm{Unif}([-1,1]^d)$.

The noise levels are sampled as $\sigma_r\sim\mathrm{Unif}([0.03,0.06])$ and $\sigma_\perp\sim\mathrm{Unif}([0.004,0.012])$. When $k_d=1$, the radius is sampled as $R_1\sim\mathrm{Unif}([2.0,3.2])$. When $k_d=2$, we sample $R_{\mathrm{out}}\sim\mathrm{Unif}([2.0,3.2])$ and $\Delta_R\sim\mathrm{Unif}([1.0,1.8])$, and set
\begin{equation}
R_1=R_{\mathrm{out}}, \qquad R_2=\max\{R_{\mathrm{out}}-\Delta_R,0.25\}.
\end{equation}
Thus $d=2$ gives one noisy circle, $d=3$ gives one noisy circle with one Gaussian coordinate, $d=4$ gives a noisy product of two circles, and $d>4$ adds $d-4$ Gaussian coordinates.

All target families used in these four scenarios have tractable oracle log densities and scores, which allows a convenient evaluation pipeline for comparing methods. 

\section{Additional Results}
\label{app:extra_dim_scaling}

Table~\ref{tab:ds_nll_per_dim_all_dim} provides additional comparison of different methods in the cross-section of $n=4,096$ with more numerical details.
Each cell in the table reports two numbers for an experimental setting: the average NLL and the corresponding standard deviation.
In general, the table delivers similar messages as in Figure~\ref{fig:ss_free_y}.
A notable observation from this table is that the NNKDE methods tend to exhibit smaller performance variation than the other methods in most high-dimensional cases.

\begin{figure*}[t]
\centering
\includegraphics[width=\textwidth]{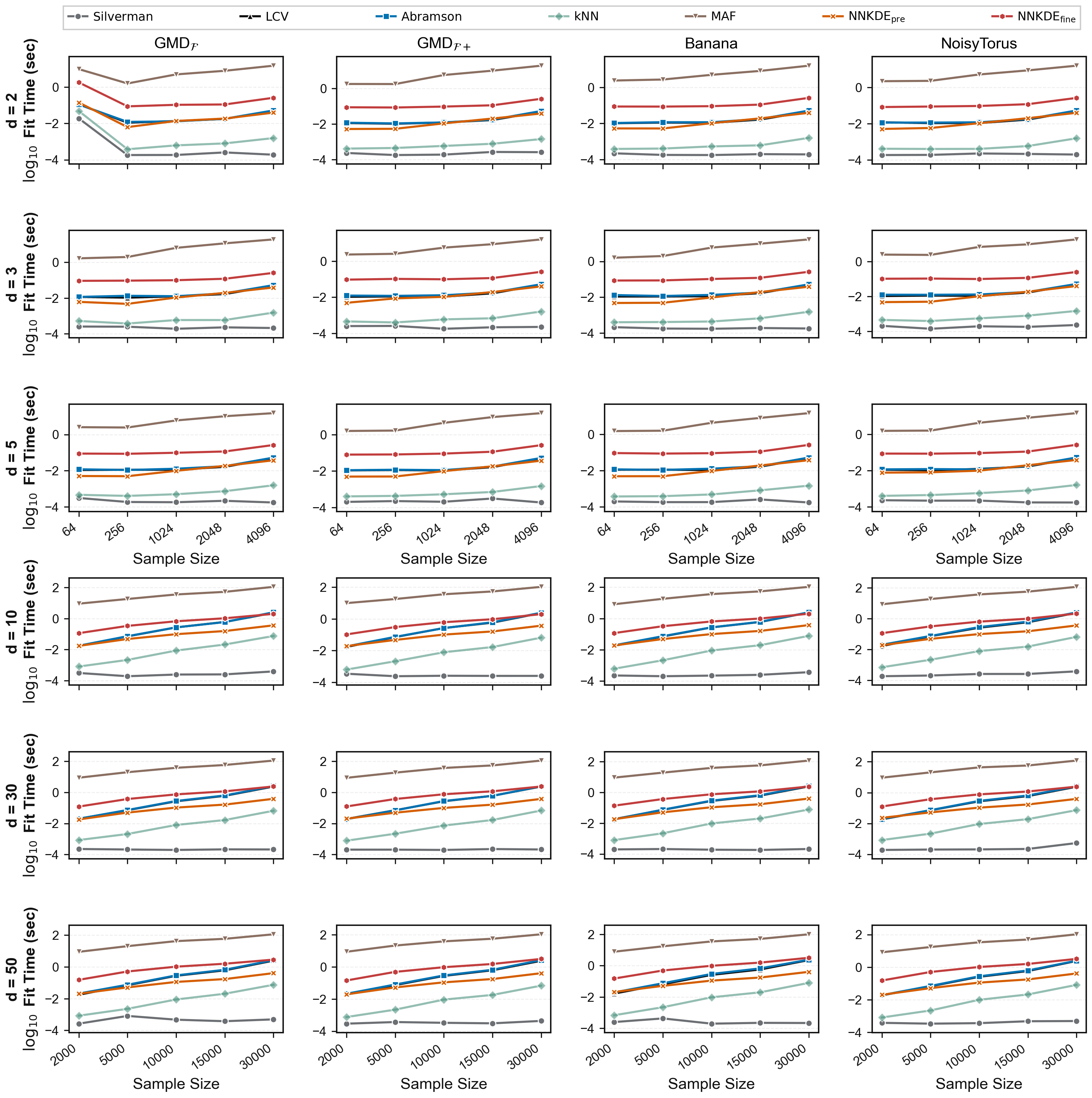}
\caption{Fitting time comparison of different methods in simulation experiments. The vertical axis reports running time in seconds on the $\log_{10}$ scale.}
\label{fig:fit_time_ss_log10}
\end{figure*}

Figure~\ref{fig:fit_time_ss_log10} compares the running time of different methods in seconds on the $\log_{10}$ scale.
The running time of the NNKDE methods covers the KDE prediction stage and the fine-tuning stage, without considering the pre-training stage.
From the Figure we can see that the proposed NNKDE methods are typically one hundred times faster than MAF, because they do not rely on freshly prepared neural networks as in MAF.
    
\begin{table*}[htbp]
\centering
\tiny
\setlength{\tabcolsep}{1.8pt}
\renewcommand{\arraystretch}{0.8}
\caption{Additional comparison of different methods in the cross-section of $n=4096$. Each entry reports mean (std) of NLL$/d$. Oracle is reported as a reference and excluded from method ranking.}
\label{tab:ds_nll_per_dim_all_dim}
\providecommand{\best}[1]{\textbf{#1}}
\providecommand{\second}[1]{\underline{#1}}

\resizebox{\textwidth}{!}{%
\begin{tabular}{@{}lccccccc@{}}
\toprule
\multicolumn{8}{c}{\textbf{Gaussian mixture distributions from the pre-training family $\cF$ (scenario GMD$_\cF$)}} \\
\midrule
Method & $d=2$ & $d=3$ & $d=5$ & $d=10$ & $d=20$ & $d=30$ & $d=50$ \\
\midrule
Silverman
& 2.166 (0.167)
& 2.137 (0.213)
& 1.872 (0.311)
& 1.944 (0.228)
& 1.978 (0.256)
& 2.050 (0.292)
& 2.211 (0.237) \\

LCV
& 2.079 (0.126)
& 1.995 (0.174)
& 1.710 (0.199)
& 1.694 (0.120)
& 1.717 (0.104)
& 1.734 (0.127)
& 1.830 (0.067) \\

$k$NN
& 2.086 (0.121)
& 2.058 (0.191)
& 1.881 (0.212)
& 2.211 (0.134)
& 2.599 (0.079)
& 2.831 (0.093)
& 3.167 (0.027) \\

Abramson
& 2.088 (0.130)
& 1.999 (0.175)
& 1.714 (0.193)
& 1.719 (0.096)
& 1.717 (0.104)
& 1.734 (0.127)
& 1.830 (0.067) \\

\midrule
MAF
& 2.197 (0.161)
& 2.208 (0.234)
& 1.857 (0.325)
& 1.803 (0.260)
& 1.720 (0.163)
& 1.689 (0.188)
& 1.733 (0.086) \\

\midrule
NNKDE$_{\text{scratch}}$
& 2.149 (0.114)
& 2.072 (0.159)
& 1.790 (0.142)
& 1.689 (0.074)
& 1.672 (0.050)
& 1.678 (0.064)
& 1.784 (0.047) \\

NNKDE$_{\text{pre}}$
& \best{2.068 (0.118)}
& \best{1.973 (0.165)}
& \best{1.664 (0.167)}
& \best{1.621 (0.089)}
& \best{1.599 (0.052)}
& \best{1.611 (0.061)}
& \second{1.687 (0.033)} \\

NNKDE$_{\mathrm{fine}}$
& \second{2.069 (0.118)}
& \second{1.976 (0.165)}
& \second{1.672 (0.168)}
& \second{1.624 (0.090)}
& \second{1.599 (0.052)}
& \second{1.611 (0.061)}
& \best{1.687 (0.032)} \\

\midrule
\textcolor{gray}{Oracle}
& \textcolor{gray}{2.056 (0.117)}
& \textcolor{gray}{1.954 (0.162)}
& \textcolor{gray}{1.640 (0.165)}
& \textcolor{gray}{1.580 (0.086)}
& \textcolor{gray}{1.545 (0.052)}
& \textcolor{gray}{1.515 (0.057)}
& \textcolor{gray}{1.538 (0.040)} \\
\bottomrule
\end{tabular}%
}

\vspace{0.85em}

\resizebox{\textwidth}{!}{%
\begin{tabular}{@{}lccccccc@{}}
\toprule
\multicolumn{8}{c}{\textbf{Gaussian mixture distributions out of the pre-training family $\cF$ (scenario GMD$_{\cF+}$)}} \\
\midrule
Method & $d=2$ & $d=3$ & $d=5$ & $d=10$ & $d=20$ & $d=30$ & $d=50$ \\
\midrule
Silverman
& 1.729 (0.266)
& 1.237 (0.448)
& 1.328 (0.466)
& 1.645 (0.409)
& 1.928 (0.397)
& 2.040 (0.340)
& 2.149 (0.186) \\

LCV
& 1.370 (0.226)
& 1.009 (0.271)
& 0.983 (0.246)
& 1.146 (0.252)
& 1.355 (0.244)
& 1.429 (0.274)
& 1.511 (0.165) \\

$k$NN
& \second{1.245 (0.205)}
& 0.970 (0.215)
& 0.984 (0.175)
& 1.310 (0.112)
& 1.729 (0.068)
& 1.940 (0.053)
& 2.231 (0.028) \\

Abramson
& 1.385 (0.228)
& 1.026 (0.285)
& 1.008 (0.251)
& 1.159 (0.230)
& 1.365 (0.217)
& 1.429 (0.274)
& 1.511 (0.165) \\

\midrule
MAF
& 1.607 (0.422)
& 1.213 (0.386)
& 1.092 (0.340)
& 0.953 (0.142)
& 1.089 (0.205)
& 1.027 (0.213)
& \second{0.969 (0.136)} \\

\midrule
NNKDE$_{\text{scratch}}$
& 1.700 (0.182)
& 1.437 (0.168)
& 1.341 (0.113)
& 1.189 (0.052)
& 1.107 (0.024)
& 1.166 (0.017)
& 1.124 (0.007) \\

NNKDE$_{\text{pre}}$
& 1.249 (0.200)
& \second{0.947 (0.196)}
& \second{0.799 (0.117)}
& \second{0.845 (0.048)}
& \second{0.802 (0.027)}
& \second{0.866 (0.019)}
& 0.981 (0.008) \\

NNKDE$_{\text{fine}}$
& \best{1.243 (0.200)}
& \best{0.940 (0.195)}
& \best{0.792 (0.117)}
& \best{0.768 (0.060)}
& \best{0.735 (0.031)}
& \best{0.753 (0.025)}
& \best{0.772 (0.011)} \\

\midrule
\textcolor{gray}{Oracle}
& \textcolor{gray}{1.210 (0.194)}
& \textcolor{gray}{0.903 (0.184)}
& \textcolor{gray}{0.764 (0.110)}
& \textcolor{gray}{0.706 (0.054)}
& \textcolor{gray}{0.664 (0.029)}
& \textcolor{gray}{0.643 (0.019)}
& \textcolor{gray}{0.632 (0.008)} \\
\bottomrule
\end{tabular}%
}

\vspace{0.85em}

\resizebox{\textwidth}{!}{%
\begin{tabular}{@{}lccccccc@{}}
\toprule
\multicolumn{8}{c}{\textbf{Banana-shaped distributions (scenario Banana)}} \\
\midrule
Method & $d=2$ & $d=3$ & $d=5$ & $d=10$ & $d=20$ & $d=30$ & $d=50$ \\
\midrule
Silverman
& -0.428 (0.326)
& -0.536 (0.293)
& -0.437 (0.179)
& -0.364 (0.133)
& -0.265 (0.103)
& -0.200 (0.067)
& -0.138 (0.049) \\

LCV
& \second{-0.429 (0.327)}
& \second{-0.537 (0.293)}
& \second{-0.438 (0.179)}
& -0.371 (0.132)
& -0.282 (0.101)
& \second{-0.227 (0.066)}
& \second{-0.181 (0.048)} \\

$k$NN
& -0.422 (0.326)
& -0.503 (0.288)
& -0.271 (0.170)
& 0.127 (0.128)
& 0.620 (0.096)
& 0.907 (0.063)
& 1.232 (0.047) \\

Abramson
& -0.424 (0.328)
& -0.530 (0.294)
& -0.421 (0.179)
& -0.235 (0.130)
& -0.121 (0.098)
& -0.077 (0.065)
& -0.043 (0.047) \\

\midrule
MAF
& \best{-0.434 (0.328)}
& \best{-0.547 (0.294)}
& \best{-0.465 (0.179)}
& \second{-0.380 (0.010)}
& \best{-0.403 (0.105)}
& \best{-0.382 (0.071)}
& \best{-0.398 (0.055)} \\

\midrule
NNKDE$_{\text{scratch}}$
& 0.977 (0.020)
& 1.023 (0.014)
& 1.059 (0.009)
& 0.945 (0.009)
& 0.883 (0.006)
& 0.995 (0.005)
& 0.955 (0.002) \\

NNKDE$_{\text{pre}}$
& -0.162 (0.189)
& -0.198 (0.154)
& 0.080 (0.046)
& 0.494 (0.027)
& 0.392 (0.013)
& 0.575 (0.009)
& 0.771 (0.004) \\

NNKDE$_{\mathrm{fine}}$
& -0.416 (0.324)
& -0.519 (0.291)
& -0.428 (0.177)
& \best{-0.399 (0.136)}
& \second{-0.284 (0.097)}
& -0.216 (0.074)
& -0.173 (0.047) \\

\midrule
\textcolor{gray}{Oracle}
& \textcolor{gray}{-0.435 (0.328)}
& \textcolor{gray}{-0.549 (0.294)}
& \textcolor{gray}{-0.469 (0.179)}
& \textcolor{gray}{-0.449 (0.136)}
& \textcolor{gray}{-0.418 (0.104)}
& \textcolor{gray}{-0.403 (0.072)}
& \textcolor{gray}{-0.424 (0.057)} \\
\bottomrule
\end{tabular}%
}

\vspace{0.85em}

\resizebox{\textwidth}{!}{%
\begin{tabular}{@{}lccccccc@{}}
\toprule
\multicolumn{8}{c}{\textbf{Noisy torus distributions (scenario NoisyTorus)}} \\
\midrule
Method & $d=2$ & $d=3$ & $d=5$ & $d=10$ & $d=20$ & $d=30$ & $d=50$ \\
\midrule
Silverman
& 1.437 (0.090)
& 1.004 (0.109)
& 1.047 (0.121)
& 0.564 (0.215)
& 0.185 (0.157)
& 0.006 (0.160)
& -0.179 (0.211) \\

LCV
& 1.107 (0.087)
& 0.553 (0.105)
& 0.650 (0.116)
& 0.022 (0.210)
& -0.431 (0.154)
& -0.634 (0.159)
& -0.838 (0.208) \\

$k$NN
& \second{0.897 (0.081)}
& 0.302 (0.087)
& 1.345 (0.135)
& 1.009 (0.254)
& 0.734 (0.245)
& 0.681 (0.193)
& 0.623 (0.285) \\

Abramson
& 1.107 (0.087)
& 0.554 (0.105)
& 0.654 (0.116)
& 0.026 (0.212)
& -0.430 (0.157)
& -0.637 (0.160)
& -0.854 (0.213) \\

\midrule
MAF
& 0.998 (0.121)
& \second{-0.039 (0.093)}
& \second{0.568 (0.118)}
& \best{-0.705 (0.368)}
& \best{-1.635 (0.111)}
& \best{-2.004 (0.158)}
& \best{-2.338 (0.143)} \\

\midrule
NNKDE$_{\text{scratch}}$
& 1.844 (0.048)
& 1.594 (0.044)
& 1.555 (0.059)
& 1.201 (0.057)
& 0.988 (0.026)
& 1.049 (0.014)
& 0.974 (0.013) \\

NNKDE$_{\text{pre}}$
& 1.294 (0.056)
& 0.743 (0.061)
& 0.957 (0.052)
& 0.794 (0.048)
& 0.500 (0.035)
& 0.657 (0.011)
& 0.841 (0.003) \\

NNKDE$_{\mathrm{fine}}$
& \best{0.604 (0.111)}
& \best{-0.420 (0.108)}
& \best{0.214 (0.071)}
& \second{-0.591 (0.223)}
& \second{-1.544 (0.063)}
& \second{-1.557 (0.046)}
& \second{-1.490 (0.013)} \\

\midrule
\textcolor{gray}{Oracle}
& \textcolor{gray}{0.498 (0.118)}
& \textcolor{gray}{-0.837 (0.119)}
& \textcolor{gray}{-0.387 (0.064)}
& \textcolor{gray}{-1.907 (0.233)}
& \textcolor{gray}{-2.788 (0.192)}
& \textcolor{gray}{-3.026 (0.216)}
& \textcolor{gray}{-3.191 (0.250)} \\
\bottomrule
\end{tabular}%
}
\end{table*}

\end{document}